\documentclass[sigconf]{acmart}

\AtBeginDocument{%
	\providecommand\BibTeX{{%
			\normalfont B\kern-0.5em{\scshape i\kern-0.25em b}\kern-0.8em\TeX}}}

\usepackage[switch,modulo]{lineno}
\usepackage{booktabs} %
\usepackage{array}
\usepackage{adjustbox}
\usepackage{tikz}
\usetikzlibrary{positioning}
\usepackage[
  separate-uncertainty = true,
  multi-part-units = repeat
]{siunitx}

\usepackage{amsmath}
\usepackage{algorithm}
\usepackage[noend]{algpseudocode}
\usepackage{amsmath}

\begin{document}
\title{A Supervised Learning Approach for Robust Health Monitoring using Face
Videos}

\author{Mayank Gupta}
\affiliation{%
	\institution{Purdue University}
}
\email{gupta369@purdue.edu}
\author{Lingjun Chen}
\affiliation{%
	\institution{Purdue University}
}
\email{chen1635@purdue.edu}
\author{Denny Yu}
\affiliation{%
	\institution{Purdue University}
}
\email{dennyyu@purdue.edu}
\author{Vaneet Aggarwal}
\affiliation{%
	\institution{Purdue University}
}
\email{vaneet@purdue.edu}
\renewcommand{\shortauthors}{M. Gupta, L. Chen, D. Yu, and V. Aggarwal}

\begin{abstract}

Monitoring of cardiovascular activity is highly desired and can enable novel applications in diagnosing potential cardiovascular diseases and maintaining an individual's well-being. Currently, such vital signs are measured using intrusive contact devices such as an electrocardiogram (ECG), chest straps, and pulse oximeters that require the patient or the health provider to manually implement. User engagement and compliance with wearables is a well-known problem that presents a significant barrier to capturing the continuous measurements needed for health monitoring. Non-contact, device-free human sensing methods can eliminate the need for specialized heart and blood pressure monitoring equipment. Non-contact methods can have additional advantages since they are scalable with any environment where video can be captured, can be used for continuous measurements, and can be used on patients with varying levels of dexterity and independence, from people with physical impairments to infants (e.g., baby camera). In this paper, we used a non-contact method that only requires face videos recorded using commercially-available webcams. These videos were exploited to predict the health attributes like pulse rate and variance in pulse rate. The proposed approach used facial recognition to detect the face in each frame of the video using facial landmarks, followed by supervised learning using deep neural networks to train the machine learning model. The videos captured subjects performing different physical activities that result in varying cardiovascular responses. The proposed method did not require training data from every individual and thus the prediction can be obtained for the new individuals for which there is no prior data; critical in approach generalization. The approach was also evaluated on a dataset of people with different ethnicity. The proposed approach had less than a 4.6\% error in predicting the pulse rate.

\end{abstract}

\maketitle
\section{Introduction}\label{sec:intro}

Regular and non-invasive measurement of vital physiological attributes such as pulse rate (PR),  pulse rate variability (PRV), and blood pressure (BP) are important due to their fundamental role in tracking one's fitness level, diagnosis of cardiovascular diseases, and monitoring of well-being. In-office and home environments, passive non-contact measurements are essential to monitor warning signs for cardiovascular diseases, stress, and anxiety. This paper explores the use of facial features from the videos to predict these vital health attributes.

Currently, the gold standard techniques for measuring such vital health attributes include using intrusive contact devices such as an electrocardiogram (ECG), chest straps, and pulse oximeters. Traditionally, ECG was extensively used for such measurement, but the recent trend has been shifted towards using pulse oximeters because of its low cost. 

Although pulse oximeters are easy to use, they have limitations for frequent measurements. First, it requires the purchase of equipment and needs either the health provider or the user to manually perform the measurements. Second, the device needs to be carried to the different places that the user goes, limiting its use. Third, the finger clip-on and earlobe clip-on may not always fit well on every individual due to the varying size of fingers and earlobes. The improper fitting of the device may lead to estimation errors \cite{haynes2007ear}. Fourth, using clip-on may be potentially uncomfortable during long use. Thus, this paper considers the use of a non-contact based approach where the passive video of the face can be used to estimate the health metrics. 

Monitoring of health parameters using non-contact methods like videos from commercially-available camera has been recently considered in \cite{verkruysse2008remote,poh2011advancements,sun2011motion,kumar2015distanceppg}, showing that the photoplethysmogram (PPG) signal can be extracted from the videos of the face.These techniques required no dedicated source of light, and a  low-cost digital camera can be used. The non-contact measurement using camera video has many applications including determination of health parameters of people working in an office environment, shop-floor, newborn infants in the hospital where using contact probes may not be possible. The non-contact method can also replace current contact methods deployed on treadmills for measurement of pulse rate.  In these works, the PPG signal is extracted from each individual, and thus the coefficients of the video features that provide the PPG signal are dependent on the individual. In contrast, we do not consider individual characteristics in the prediction. The proposed method can thus help predict health metrics of an individual for which no training sample has been collected in the past, making our methodology robust. Also, the use of non-contact methodology has an additional advantage of being scalable, and portable since cameras are ubiquitous. 

The proposed approach has two key steps. The first step considers capturing the video and extracting the face from the video. The features corresponding to the face in each frame are obtained. The second step includes training a  neural network to learn the health parameters from the above obtained features. Our results obtained from the deep learning model has a mean absolute percentage error of 4.6\% for predicting pulse rate. The appendix contains initial results on predicting indicators of the variance in pulse rate.

\section{Related Work}\label{sec:related}
 
The authors in \cite{humphreys2007noncontact} showed that it is possible to extract the PPG signal from the video using a complementary metal-oxide semiconductor camera by illuminating a region of tissue using through external light-emitting diodes at dual-wavelength (760nm and 880nm).  Further, the authors of  \cite{verkruysse2008remote} demonstrated that the PPG signal can be estimated by just using ambient light as a source of illumination along with a simple digital camera.  Further in \cite{poh2011advancements}, the PPG waveform was estimated from the videos recorded using a low-cost webcam. The red, green, and blue channels of the images were decomposed into independent sources using independent component analysis. One of the independent sources was selected to estimate PPG and further calculate HR, and HRV. All these works showed the possibility of extracting PPG signals from the videos and proved the similarity of this signal with the one obtained using a contact device. Further, the authors in \cite{10.1109/CVPR.2013.440} showed that heart rate can be extracted from features from the head as well by capturing the subtle head movements that happen due to blood flow.

The authors of \cite{kumar2015distanceppg} proposed a methodology that overcomes a challenge in extracting PPG for people with darker skin tones. The challenge due to slight movement and low lighting conditions during recording a video was also addressed. They implemented the method where PPG signal is extracted from different regions of the face and signal from each region is combined using their weighted average making weights different for different people depending on their skin color. 

There are other attempts where authors of \cite{6523142,6909939, 7410772, 7412627} have introduced different methodologies to make algorithms for estimating pulse rate robust to illumination variation and motion of the subjects. The paper \cite{6523142} introduces a chrominance-based method to reduce the effect of motion in estimating pulse rate. The authors of \cite{6909939} used a technique in which face tracking and normalized least square adaptive filtering is used to counter the effects of variations due to illumination and subject movement. 
The paper \cite{7410772} resolves the issue of subject movement by choosing the rectangular ROI's on the face relative to the facial landmarks and facial landmarks are tracked in the video using pose-free facial landmark fitting tracker discussed in \cite{yu2016face} followed by the removal of noise due to illumination to extract noise-free PPG signal for estimating pulse rate. 

Recently, the use of machine learning in the prediction of health parameters have gained attention. The paper \cite{osman2015supervised} used a supervised learning methodology to predict the pulse rate from the videos taken from any off-the-shelf camera. Their model showed the possibility of using machine learning methods to estimate the pulse rate. However, our method outperforms their results when the root mean squared error of the predicted pulse rate is compared. The authors in \cite{hsu2017deep} proposed a deep learning methodology to predict the pulse rate from the facial videos. The researchers trained a convolutional neural network (CNN) on the images generated using Short-Time Fourier Transform (STFT) applied on the R, G, \& B channels from the facial region of interests.
The authors of \cite{osman2015supervised, hsu2017deep} only predicted pulse rate, and we extended our work in predicting variance in the pulse rate measurements as well.

All the related work discussed above utilizes filtering and digital signal processing to extract PPG signals from the video which is further used to estimate the PR and PRV.  %
The method proposed in \cite{kumar2015distanceppg} is person dependent since the weights will be different for people with different skin tone. In contrast, we propose a deep learning model to predict the PR which is independent of the person who is being trained. Thus, the model would work even if there is no prior training model built for that individual and hence, making our model robust. 

\section{Data Collection}\label{sec:data}
We designed our own experiment to collect the data for training the model. Twenty healthy volunteers participated in this study. The participants were recruited from a university population through email including a description of the study. This study was reviewed by the university's Institutional Review Board, and all participants provided informed consent. The details of our experiment are given below:
\subsection{Set-Up}
To predict the vital health metrics, we used the face video of the person. The video is obtained using a 5 MP front-facing Hello face-authentication camera (1080p HD) from Microsoft Surface Book, having 30 frames per second.  %
The camera is capable of capturing red, green and blue color channels. The authors of \cite{verkruysse2008remote} suggested that green channel of the video outperforms blue and red channel in estimating the health parameters. Therefore, we also utilize the green channel of the camera. The features obtained from the video will be used to predict the health metrics. 

To train the data, true values of PR is calculated using a contact measurement device, Shimmer3 GSR+, that records the ground truth PPG signal. We chose earlobe as the suitable position for recording PPG since it is close to the face. We care about this proximity since face is used for our video recordings as well. Subjects were asked to sit still for a 50s video, facing towards the camera at a distance of approximately 0.5m, and the PPG signal was recorded simultaneously through the Shimmer3 GSR+ device. The experimental set-up for conducting the experiments is shown in Figure \ref{fig:experiment} in appendix.  

\begin{figure*}[htbp]
	\centering
	\includegraphics[width=\textwidth]{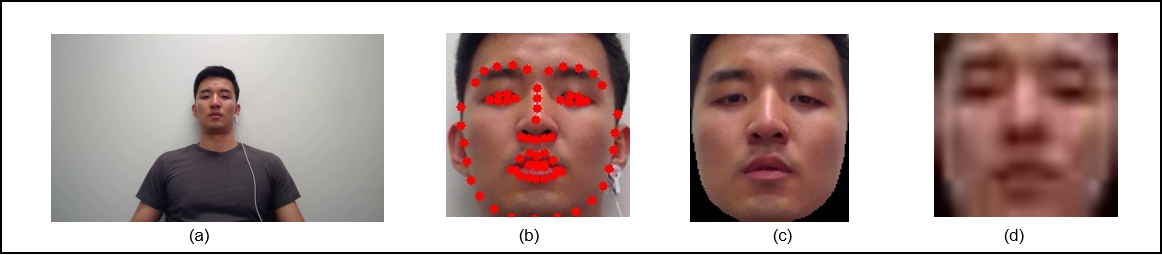}
	
	\caption{The steps followed for feature extraction from each frame of the video. (a) The actual image (one of the many frame) from the video captured during the experiment. (b) The detected and aligned face using DeepFace along with landmark points. (c) The face is cropped using landmark points to get only required face features. (d) Each frame is downsampled to 20x20 image}
		\vspace{-.2in}
	
	\label{fig:process}
\end{figure*}

\subsection{Experiments}

The study involved people of different skin colors and ethnicities. For each subject, the measurements were performed after different activity levels, thus providing a variation in the heart rate of each subject. The different activity levels at which the measurements were collected were:  Rest Position, Brisk Walk, and Exercise.

\textbf{Rest Position}: The first experiment was conducted when each participant was in rest condition. Each subject was asked to relax and sit in front of the camera. The video and Shimmer device recordings were collected simultaneously.

\textbf{Brisk Walk}: The next experiment involved data collection of the same participants after they were asked to do a brisk walk for 0.25 miles at a speed of 3-5 mph on the treadmill.  The video and shimmer recordings were captured immediately after the subject complete the brisk walk. 

\textbf{Exercise}:  The last experiment involved more challenging physical tasks. All the subjects were asked to perform as many push-ups or sit-ups as they can such that they exert themselves to their full capacity. This activity was designed to elicit a high pulse rate since the individual was working out at their full capacity.

The mean pulse rate for rest, walk, and exercise conditions were 72.9, 79.6, and 98.5 respectively.  PPG and facial data were videos recorded immediately after each activity to minimize recovery effects on the physiological data. Each subject was given rest of 10 minutes before each activity so they can recover prior to the next activity.

We acknowledge that the heart rate will dynamically change during the video capture; however, the purpose of this study was to compare device-free sensing to the gold standard continuous measurement. The success of capturing these dynamic behaviors may further show the promising capability of the technique in addressing the complexity of dynamic changes commonly seen in the real world.
\section{Proposed Approach}\label{sec:method}
The methodology adopted for the estimation of health parameters is two-fold.
The videos of subjects captured under different conditions were processed to extract the face features as the first step. The next step involves training the deep learning model using the face features as predictors and actual values of PR as the response variables. The detailed steps are explained below.
\subsection{ Video Processing}
The video was recorded for 50 seconds for each subject under different activities. The entire video was broken into frames where each frame was composed of red, green, and blue color bands. We utilized the DeepFace \cite{taigman2014deepface} algorithm for the purpose of face recognition. DeepFace algorithm was developed by the researchers at Facebook and had an accuracy of 97.35 \% on the \textit{Labeled Faces in the Wild} (LFW) \textit{dataset}, which reduced the error of the current state of the art \cite{huang2012learning,sun2013deep,cao2013practical,chen2013blessing} by more than 27\%. DeepFace utilized a nine-layer deep neural network and was trained on a large facial dataset of four million facial images belonging to more than 4,000 identities.

The first step in video processing involved the detection of human faces in each frame of the video. The detected human face was then aligned automatically by DeepFace using the 3D alignment method \cite{taigman2014deepface}. The aligned face was cropped from the image using the landmark points on the face shown in Figure \ref{fig:process}. The image was cropped  to only facial features and removing extra pixel values from the images. We were careful in retaining the forehead since it contained the maximum information about blood perfusion inside the arteries \cite{kumar2015distanceppg}. %
  The cropped images were used for training the deep learning model. The stages of video processing are shown in Figure \ref{fig:process}.   

  The extraction of ''right'' features is important as it plays a significant role in training a neural network.  Choosing the subset of features from the available data reduces redundancy in the input to the neural networks and subsequently improving the performance. Therefore, we down-sample each cropped image to 20x20 image and extract 400 pixels intensity values from each frame as shown in Figure \ref{fig:process} (d). 

 \subsection{Model Training}
We trained a deep learning model using TensorFlow to estimate health metrics. The model was trained through a multi-layered neural network.   We used a fully connected neural network with three hidden layers. The detailed architecture of the network used in shown in Figure \ref{fig:arch} (in Appendix). The network consisted of rectified linear units (ReLU) and the rectifier activation was given  as $f(x)=max(0,x)$, where $x$ was the input to the neuron. The choice of network architecture and activation function was dependent on the minimum value of the loss function. The network was trained using a backpropagation algorithm with the mean squared error as the loss function.  Batch normalization was used in each hidden layer \cite{DBLP:journals/corr/IoffeS15}. The use of drop-out was one of the simplest ways to avoid over-fitting of the neural network \cite{srivastava2014dropout}. The drop-out rate was set to 30\% to avoid over-fitting in all the three hidden layers. This will help in better generalizing the network for unseen data. The green color band was shown to be the best source for extracting information about the health parameters \cite{verkruysse2008remote}. Hence, we utilized all the pixel values corresponding to the green channel of each frame to train our machine learning model. The image from each frame was downsampled to a 20x20 image. Hence, we used 400 features from each frame to train the model. The features were normalized to bring them in a range of [0,1] so that it was easier for the neural network to learn from the data. The downsampling was also done to reduce the computational expense of our model. The actual response value, i.e., PR was extracted from the PPG signal recorded during our experiment. In order to extract actual PR from PPG, we computed power spectral density (PSD) of the PPG signal using a fast fourier transform (FFT) algorithm. The PR was then estimated as the frequency corresponding to the maximum power in the PSD (PR= 60. $\mathit{f}$ bpm), where $\mathit{f}$ is the required frequency.

\section{Prediction Results}\label{result}

\begin{figure}[htbp]
	\centering
		\vspace{-.2in}
	\includegraphics[width=0.48\textwidth]{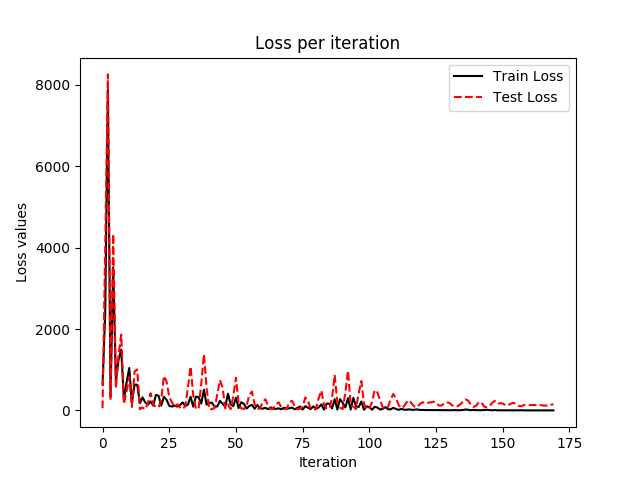}
		\vspace{-.4in}
	\caption{Behavior of train loss and test loss for predicting pulse rate.}
	\label{fig:HR_loss}
		\vspace{-.2in}
\end{figure}

\begin{figure}[htbp]
	\centering
		\vspace{-.2in}
	\includegraphics[ width=.45\textwidth]{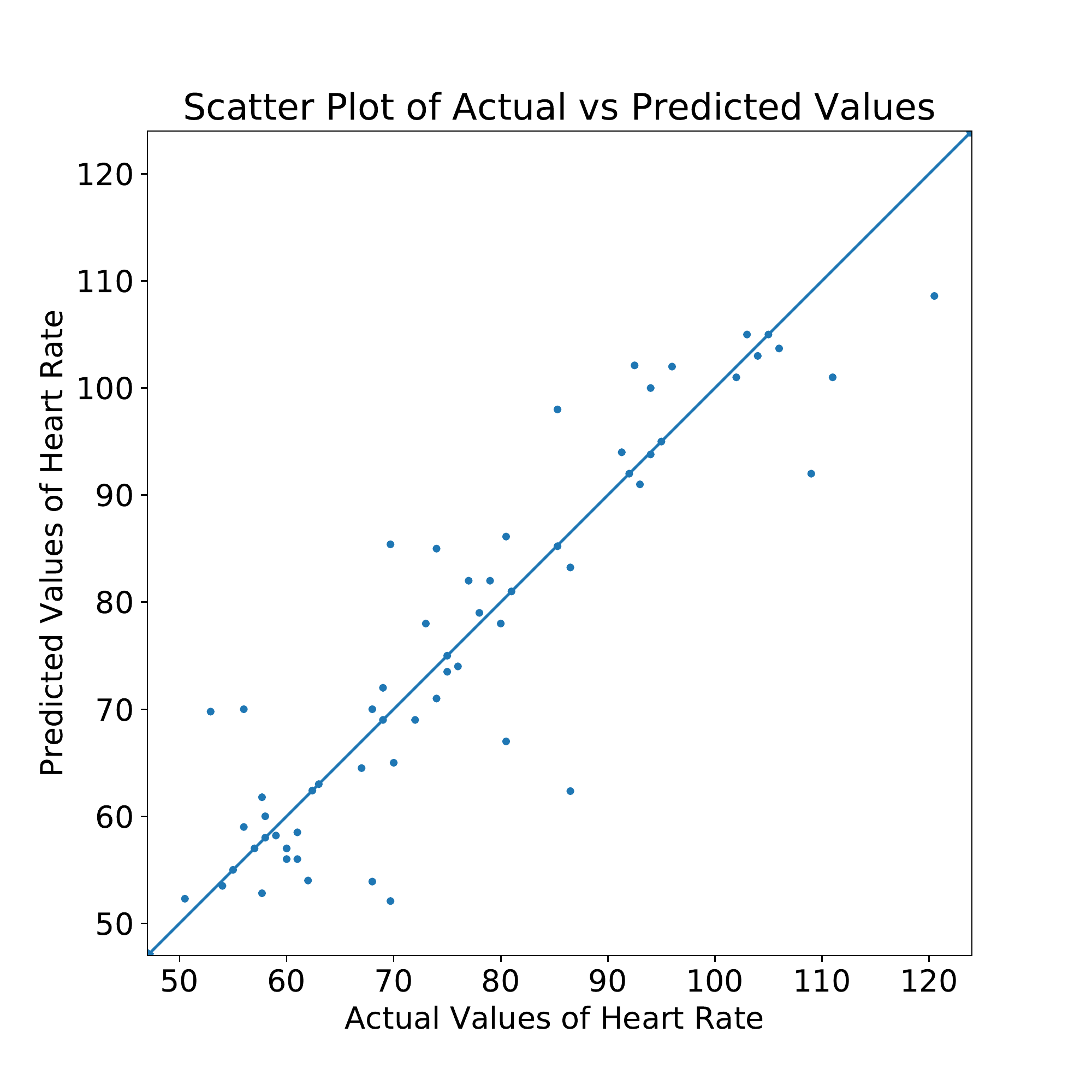}
		\vspace{-.3in}
	\caption{Scatter plot of the predicted PR value vs. the ground truth PR value. The straight line is function \(y=x\). The closeness of the points to the line indicates the model accuracy. }
	\label{fig:scatter}
	\vspace{-.2in}
\end{figure}

The values of PR predicted by our model were compared with the true values calculated from the readings of a contact device and the errors were calculated accordingly. The mean absolute percentage errors were calculated using leave-one-out cross-validation. To be more specific, since we had 20 subjects in total, we chose nineteen out of the twenty subjects and picked out all observations from those nineteen subjects to make up our entire training set. All observations for the subject that was left out from the training set were considered as the test set. We iterated this procedure for each subject to make sure that we test our model on each individual. Since the data from the test set is completely new compared to the training set, this tells us how our model predicts subjects it has never seen before, regardless of skin tone, race, and facial features.

We then calculated the mean absolute percentage error (MAPE) and Root Mean Squared Error (RMSE) for our predictions. The mean of errors on all 20 subjects was found to be 4.6\%. Similarly, the RMSE value for our test set is found out to be 4.39.  The authors of \cite{osman2015supervised} reported an RMSE of 9.52 on the test set in predicting PR meaning that our model outperforms theirs and shows a reduction in RMSE by 53\% for predicting PR.

Figure \ref{fig:HR_loss} shows how the test and train loss varies with the number of iterations run by our network. For our computation, we used mean squared error as the loss function. The number of iterations was chosen based on the behavior of test and train loss. If the number of iterations was too low, it leads to under-fitting wherein both train and test errors were high and if the number of iterations was too many, it leads to over-fitting. To avoid these scenarios, we ran our model at 170 iterations.

Figure \ref{fig:scatter} shows a scatter plot between predicted and actual values of pulse rate. The straight line shown is a 45-degree line (\(y=x\)), and the closeness of the scatter points to the straight line indicates the high accuracy of our model.   

\section{Conclusions}\label{sec:concl}

The monitoring of the health parameters like PR and PRV is important to keep a check on the individual's health and spot the potential cardiovascular diseases. Recently, the use of device-free methods such as using camera videos is preferred over contact methods like pulse oximeters for such measurement. In this paper, we proposed a two-fold methodology wherein a supervised learning technique is leveraged to predict the pulse rate. The physiological parameters are remotely predicted using the video of human faces captured using a laptop's camera. The subtle changes in the face pixels intensity over the different frames of the video are exploited to train a neural network with three hidden layers. Experimental evaluations are performed for twenty subjects, and the proposed approach demonstrates significant improvement as compared to the baselines thus validating that the approach has the potential to be applied in real scenarios.

\bibliographystyle{IEEEtran}

\bibliography{bibfile}
\newpage
\clearpage
\appendix
\section{Experimental Setup and Neural Network}

\begin{figure}[htbp]
	\centering\
	\includegraphics[width=0.48\textwidth]{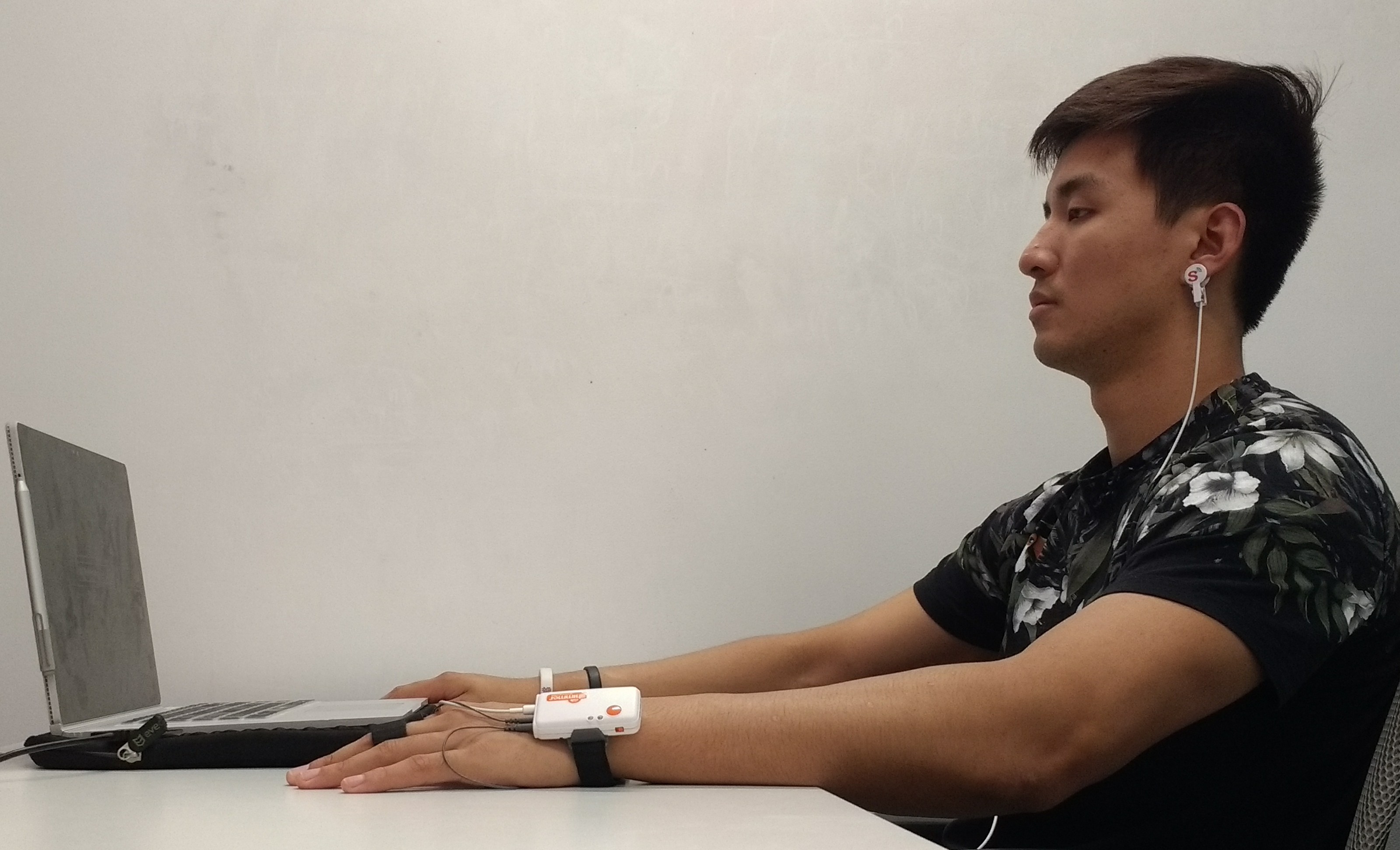}
	\vspace{-.2in}
	\caption{Experimental set-up. The contact probe of the Shimmer3 GSR+ is attached to the earlobe and laptop camera is placed around 0.5 m away from the subject.}
	\label{fig:experiment}
		\vspace{-.1in}
\end{figure}

\begin{figure}[htbp]
	\centering
	\includegraphics[width=0.48\textwidth]{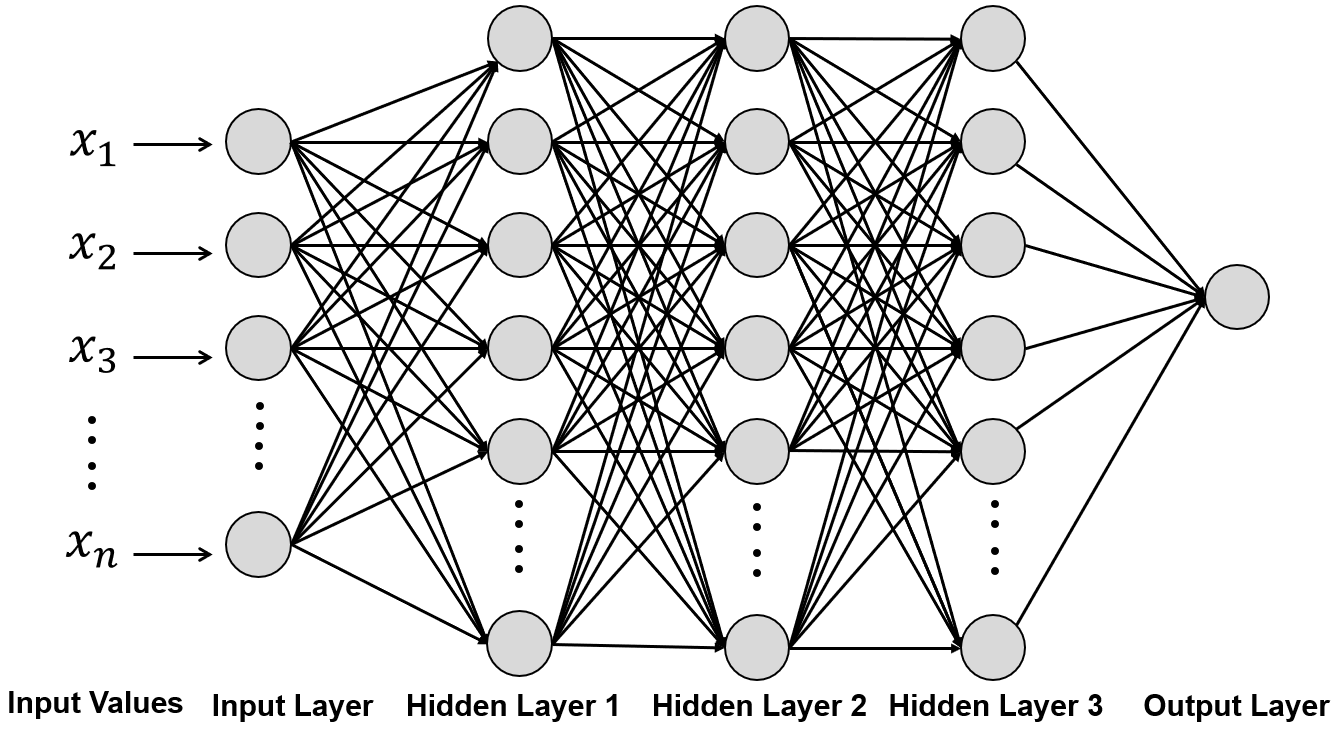}
		\vspace{-.2in}
	\caption{The architecture of a fully connected neural network with three hidden layers}
		\vspace{-.1in}
	\label{fig:arch}
\end{figure}

Figure \ref{fig:experiment} depicts the experimental setup. Further, Fig. \ref{fig:arch} depicts the used fully connected neural network architecture. 

\begin{figure}[htbp]
	\centering
	\includegraphics[trim=.2in .5in .2in .5in, clip, width=.48\textwidth]{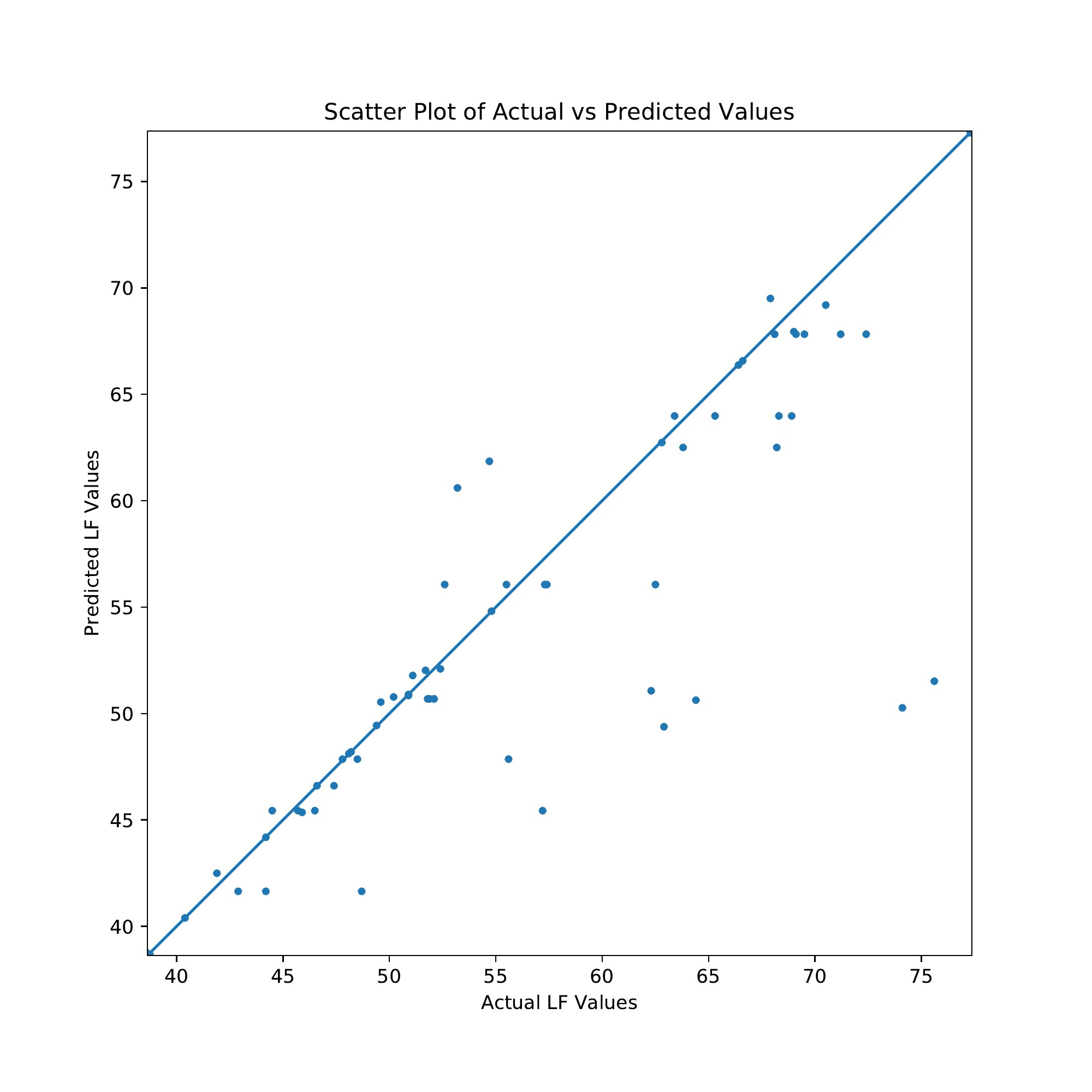}
	\vspace{-.2in}
	\caption{Scatter plot of the predicted LF value vs. the ground truth LF value.}
	\label{fig:scatter_LF}
	\vspace{-.1in}
\end{figure}
\section{Pulse Rate Variability}\label{sec:appendix}

\begin{figure}[htbp]
	\centering
	\includegraphics[width=0.48\textwidth]{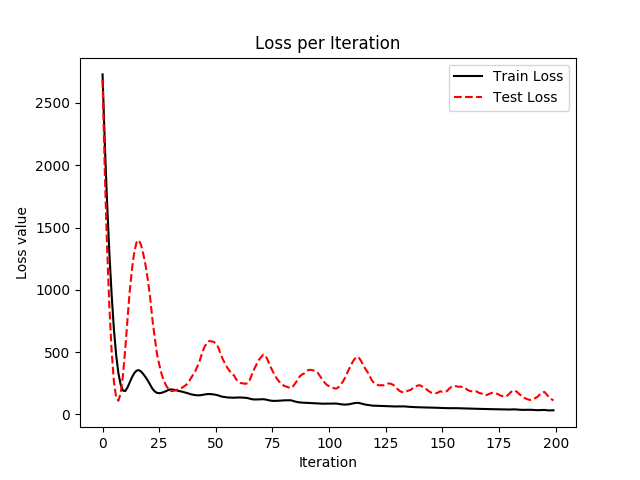}
	\vspace{-.1in}
	\caption{Training loss and test loss for predicting Low Frequency (LF) component of PRV. }
	\label{fig:LF}
	\vspace{-.1in}
\end{figure}

\begin{figure}[htbp]
	\centering
	\includegraphics[width=.485\textwidth]{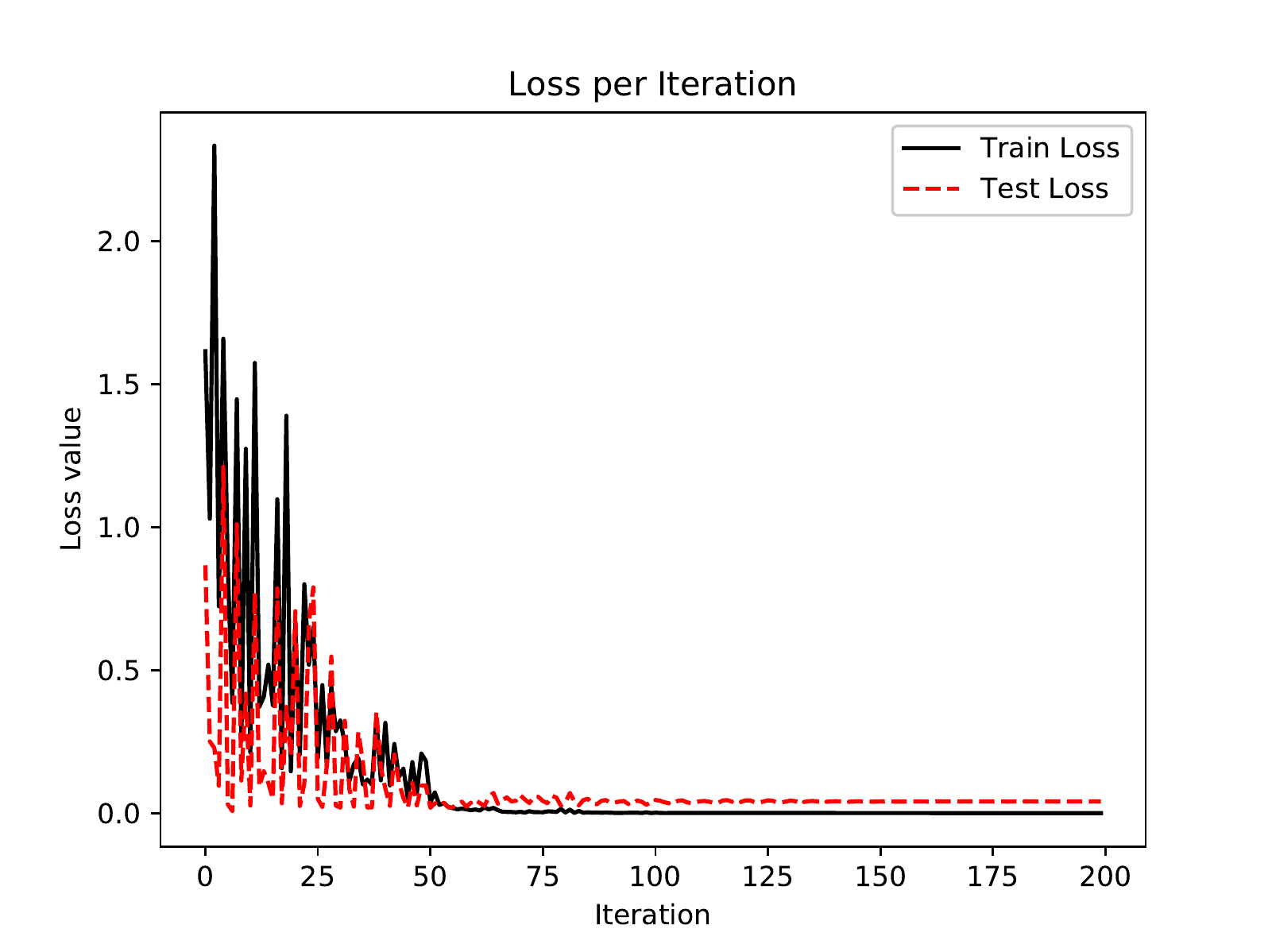}
	\vspace{-.1in}
	\caption{Training loss and test loss for predicting High Frequency (HF) component of PRV.}
	\label{fig:HF}
	\vspace{-.1in}
\end{figure}

\begin{figure}
	\centering
	\includegraphics[trim=.2in .5in .2in .5in, clip, width=.48\textwidth]{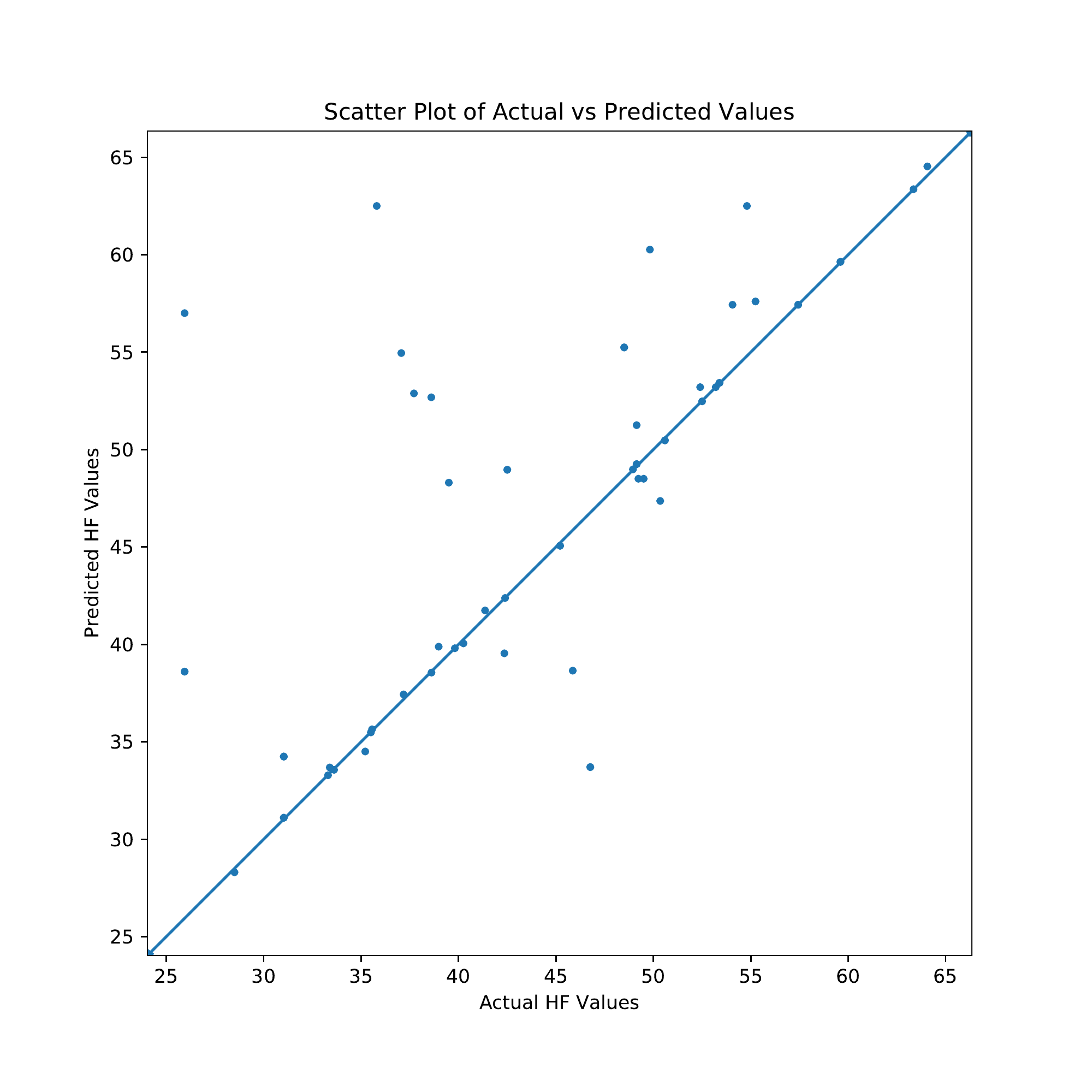}
	\vspace{-.1in}
	\caption{Scatter plot of the predicted HF value vs. the ground truth HF value.}
	\label{fig:scatter_HF}
	\vspace{-.1in}
\end{figure}

Pulse  rate  variability  is  the  variation  in  the  time  interval between two expansions of the artery. It is usually measured by the variation in beat-to-beat interval. This metric is considered as a non-invasive technique for measuring autonomic nervous system  (ANS)  activity.  The  autonomic  nervous  system has  two  branches;  sympathetic  nervous  system  (SNS)  and parasympathetic  nervous  system  (PNS)  and  is  regulated  by hypothalamus.  Its  function  includes  control  of  respiration, cardiac regulation, vasometer activity and certain reflex actions like  coughing,  sneezing,  swallowing,  and  vomiting.  High-frequency  (HF)  component  of  PRV  is  affected  by  efferent vagal  (parasympathetic)  activity  and  it  decreases  during  the conditions  of  acute  time  pressure,  emotional  strain,  mental stress,  and  elevated  anxiety.  The  low-frequency (LF) component of PRV is known to contain both sympathetic and  vagal  influences.  Thus,  frequent  and  accurate  measurement  of  PR  and  PRV  can  provide  critical  signs  of  one's well being and any abnormality could lead to potential health problems.

In this study, the LF and HF components of PRV were roughly estimated by computing the area under the PSD curve between specific frequency range. For LF, the frequency range is 0.04-0.15Hz and for HF, it is 0.15-0.4Hz. We used these rough estimates of HF, and LF as a response variable to train our model.   We trained our model using High-Frequency component and Low-Frequency component of the PPG signal. We designed separate models to train our model to predict  the HF component and the LF component. 

 We  tested our model using leave-one-out cross validation method similar to the pulse rate. Our model makes predictions on the user who has not been seen before in the training data. We choose the  number of iterations to run our model as 200. The test and training losses with iterations for the LF and the HF component of the PPG signal are depicted in  Fig. \ref{fig:LF} and Fig. \ref{fig:HF}, respectively.  

For predicting LF component, mean absolute percentage error on test set is 4.58\%, and  the root mean squared error is 3.49. On the other hand, the MAPE for  HF component is found to be 10.2\%,  and the RMSE is 4.96 for test data.  The mean RMSE for our model is 4.3 whereas the mean RMSE taken over different skin colored people in \cite{kumar2015distanceppg} is 25.3 thus providing 83\% decrease in the RMSE. Figures \ref{fig:scatter_LF} and \ref{fig:scatter_HF} depict the comparison between the actual values and the predicted values for the two components of the PPG signal, respectively. We note that dataset in \cite{kumar2015distanceppg} \& \cite{osman2015supervised} is different from the dataset used in this paper, thus not providing comparison on the same dataset. However, since the code or data of the prior work is not public, the comparison is made on the aggregate prediction accuracy.

\end{document}